\documentclass{article} 

\usepackage[table,xcdraw,dvipsnames]{xcolor}
\usepackage{graphicx}	                        
\usepackage{amsmath}	                        
\usepackage{amssymb}	
\usepackage{booktabs}                           
\usepackage{times}
\usepackage{microtype}
\usepackage{epsfig}
\usepackage{caption}
\usepackage{float}                              
\usepackage{placeins}
\usepackage{color, colortbl}
\usepackage{stfloats}
\usepackage{enumitem}
\usepackage{tabularx}                           
\usepackage{xstring}
\usepackage{multirow}
\usepackage{xspace}
\usepackage{url}
\usepackage{subcaption}
                       
\usepackage{adjustbox}                          
\usepackage{tabulary}
\usepackage{amsfonts}                           
\usepackage{bm}                                 
\usepackage{lipsum}                             
\usepackage[normalem]{ulem}                     
\usepackage{makecell}                           
\usepackage{multicol}

\usepackage{hyperref}
\hypersetup{colorlinks=true,citecolor={Violet},linkcolor={RoyalPurple},urlcolor={RoyalPurple}}
\usepackage[capitalize]{cleveref}

\makeatletter
\DeclareRobustCommand\onedot{\futurelet\@let@token\@onedot}
\def\@onedot{\ifx\@let@token.\else.\null\fi\xspace}

\def\eg{\emph{e.g}\onedot}

\makeatother

\newcommand{\citi}[1]{\footnotesize{\textit{\textbf{[1, 2, 3]}}}\normalsize}            

\newcommand{\inr}{ARC}
\newcommand{\monospace}[1]{{\fontfamily{lmtt}\selectfont #1}}

\usepackage{mathtools}

\usepackage{wrapfig}

\usepackage{iclr2025_conference,times}

\usepackage{amsmath,amsfonts,bm}

\def\eqref#1{equation~\ref{#1}}

\def\1{\bm{1}}

\DeclareMathAlphabet{\mathsfit}{\encodingdefault}{\sfdefault}{m}{sl}
\SetMathAlphabet{\mathsfit}{bold}{\encodingdefault}{\sfdefault}{bx}{n}

\title{\inr{}: Anchored Representation Clouds for High-Resolution INR Classification}

\author{Joost Luijmes,\quad Alexander Gielisse,\quad Roman Knyazhitskiy,\quad Jan van Gemert\thanks{Correspondence to joostluijmes.academia@gmail.com}\\
Computer Vision Lab\\
Delft University of Technology\\
The Netherlands\\
}

\iclrfinalcopy 
\begin{document}

\maketitle

\vspace{-1cm}

\begin{abstract}

Implicit neural representations (INRs) encode signals in neural network weights as a memory-efficient representation, decoupling sampling resolution from the associated resource costs. Current INR image classification methods are demonstrated on low-resolution data and are sensitive to image-space transformations. We attribute these issues to the global, fully-connected MLP neural network architecture encoding of current INRs, which lack mechanisms for local representation: MLPs are sensitive to absolute image location and struggle with high-frequency details. We propose \inr{}: Anchored Representation Clouds, a novel INR architecture that explicitly anchors latent vectors locally in image-space. By introducing spatial structure to the latent vectors, \inr{} captures local image data which in our testing leads to state-of-the-art implicit image classification of both low- and high-resolution images and increased robustness against image-space translation. Code can be found at {\small \url{github.com/JLuij/anchored_representation_clouds}}.

\end{abstract}
\vspace{-.4cm}

\section{Introduction}
\label{sec:intro}

From novel view synthesis to inverse problems, implicit neural representations (INRs) have enabled leaps in accuracy across a variety of problems and domains due to their unique data compression and generalisation capabilities \citep{mildenhall2021nerf, essakine2024wherestandoninrs}. As such, interest has grown in whether INRs can similarly enrich conventional computer vision tasks like image classification; the focus of this research. 

\begin{wrapfigure}{r}{0.45\textwidth}
\vspace{-.8cm}
\includegraphics[width=\linewidth]{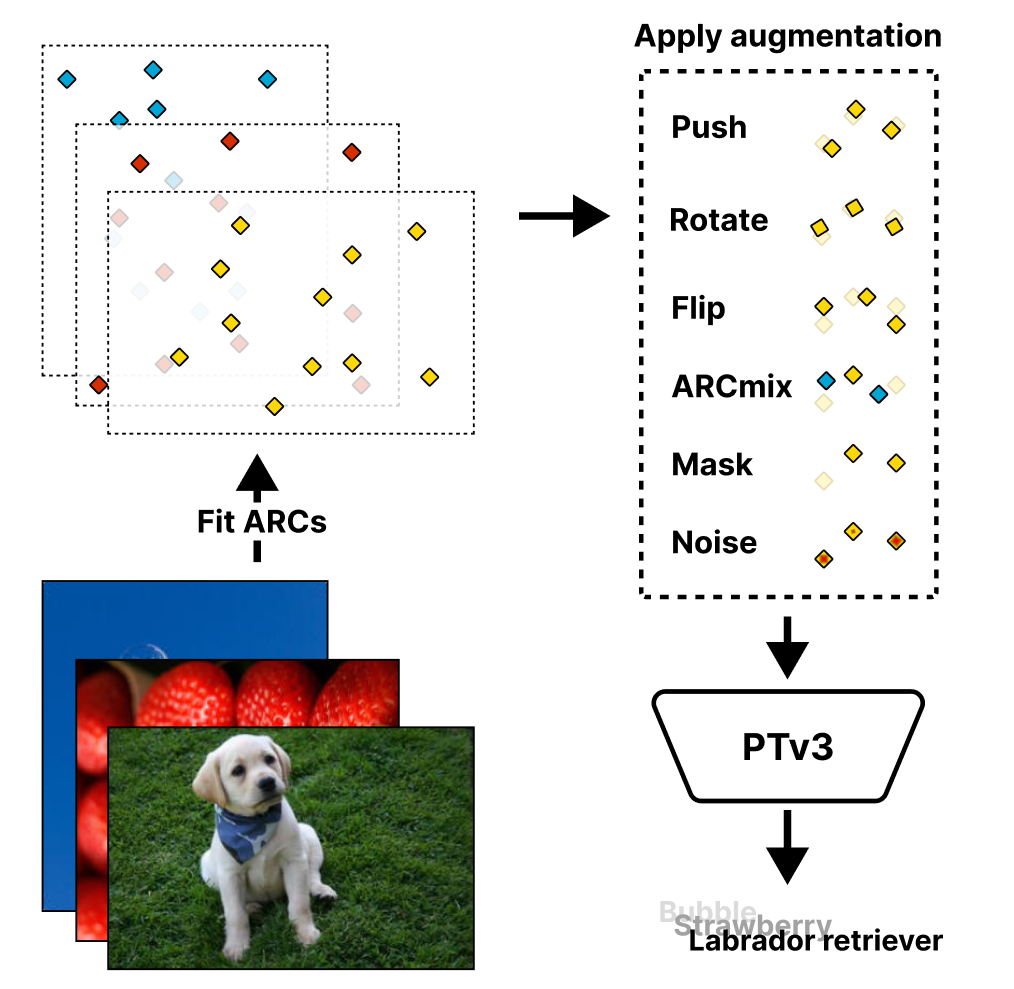}
        \caption{\inr{} anchors latent vectors directly in image coordinate space, preserving the local spatial image structure within the INR weight-space. Once trained, \inr{} can be processed by a point cloud classifier.
        } \label{fig:visualabstract}
\vspace{-.3cm}
\end{wrapfigure}


An INR is a neural network (NN) that learns a mapping from coordinates in the signal's domain to the signal's values, \eg from 2D pixel coordinates to RGB colours. After training, the INR can reconstruct an approximation of the original signal when queried on all signal coordinates, implying that the signal is encoded inside the INR weights. INRs are able to compress signals in a variety of domains \citep{dupont2021coin, strumpler2022, mod-agnostic-compr_inr_2023, fons2022hypertime, huang2022compressingweather}, and exhibit excellent generalisation capabilities outside of the signal's domain \citep{mildenhall2021nerf, yu2021pixelnerf, chen2021liif}. 

Images are formed by sensor elements placed on a grid. In a grid, elements of equal size are positioned equidistantly over the  domain, irrespective of the image content. 
This uniform data representation results in memory costs that scale exponentially with signal dimensionality and resolution. To make training on image datasets feasible, images are typically downsampled, which may eliminate relevant features \citep{hou2016patch, tinyobjectsCnn} or lead to decreased performance \citep{jiang2020defensegridfeats, tan2019efficientnet}. While these issues can be mitigated \citep{tinyobjectsCnn, hou2016patch}, INRs offer a more efficient, content-based, image representation.

Current INR classification methods \citep{kofinas2024graph_nnsaregraphs, kalogeropoulos2024scaleequivariant, zhou2024permutation} face several shortcomings. First, these methods are only demonstrated on low-resolution image datasets such as MNIST \citep{lecun1998mnist}, Fashion-MNIST \citep{fashionmnist}, and CIFAR10 \citep{krizhevsky2009cifar}. Second, the employed INR architecture learns entirely different representations under camera transformations such as translation. Translation equivariance is a fundamental property in traditional image classifiers \citep{cohenc16gcnn, zeiler2014viscnns}, especially because higher-resolution image datasets allow for less restricted object positions and present challenges in terms of background noise \citep{datasetcenterbias}. Third, INR classifiers typically suffer from overfitting \citep{kofinas2024graph_nnsaregraphs, shamsian2024inr_augment,  kalogeropoulos2024scaleequivariant}, with restricted data augmentation methods to combat this \citep{dwsnets, shamsian2024inr_augment}, resorting to the resource-intensive process of fitting several redundant INRs per image to improve generalisation \citep{dwsnets, spatialfuncta, shamsian2024inr_augment, zhou2024permutation}.

In this paper, we introduce a novel type of INR named \inr{}: Anchored Representation Clouds, along with a flexible classification pipeline and data augmentation methods (\autoref{fig:visualabstract}). An \inr{} consists of 1) an specific encoder which anchors a cloud of latent vectors in image coordinate space, and 2) an MLP decoder that is shared among \inr{} instances. In querying a coordinate, \inr{} finds the nearest latent vectors and decodes their content. This way, \inr{} retains local image features, making it more robust against image translation and more performant in higher-complexity image classification. Furthermore, as the latent vectors can be positioned freely, they can be anchored more densely in high-frequency image regions, biasing model capacity towards complex regions rather than encoding the image globally. By increasing the number of anchored latents, \inr{} can trivially scale to larger image complexity.
By converting images to a set of \inr{}s, each image is effectively represented by a cloud of latent vectors, allowing point-cloud architectures for downstream use, along with intuitive and effective point-cloud data augmentation methods which eliminate the need for expensive redundant INR fitting.
To our knowledge, this is the first work to utilise an INR's entire weight-space on a high-resolution dataset like Imagenette \citep{imagenette}, achieving a classification accuracy of $75.92\%$. We further demonstrate \inr{}’s robustness to image-space transformations and its ability to capture high-resolution images.

Our contributions include the following. (1) A novel INR architecture that anchors latent vectors in the image coordinate space, preserving spatial locality. (2) An accompanying classification pipeline that enables effective and intuitive weight-space augmentation methods. (3)  Enhanced robustness to image-space translations in INR classification.

\section{Related Work}\label{sec:related}
\textbf{Implicit neural representations.} 
An implicit neural representation (INR) \citep{sitzmann2019siren} is a neural network trained to represent a signal. In the image domain, an INR aims to learn a mapping between pixel coordinates and pixel values. After training, a forward pass of an INR produces a \textit{reconstruction} of the original image, which is now captured inside the INR's weights. Alternative names for implicit neural representations are neural fields \citep{nefs_and_beyond_2022, papa2024trainnef, wessels2024grounding}, coordinate networks \citep{martel2021acorn, lindell2022bacon, zheng2022fourfeatanalysis} and coordinate-based neural representation \citep{tancik2021learnedinits}.

The premise of INRs does not prescribe any particular architecture, prompting early work to assess the suitability of the simple MLP \citep{mildenhall2021nerf, deepsdf2019park, mescheder2019occupannetworks}. Such models struggle to capture the high-frequency components of the signal, due to the \textit{spectral bias}, which in INR-context results in blurry image reconstructions \citep{rahaman19_nnspecbias, tancik2020fourfeat, sitzmann2019siren}. This issue inspired a large set of diverse solutions, such as transforming the input using sinusoids \citep{tancik2020fourfeat, mildenhall2021nerf, zheng2022fourfeatanalysis}, modifying the activation function \citep{sitzmann2019siren, ramasinghe2022unifying_inr_activations, gaussianactivations, waveletactivations, liu2024finer, gao2024hsiren}, or positioning learnable elements in the signal coordinate space to represent local image regions \citep{liu2020sparsevoxfields, peng2020convoccupancynets, jiang2020local, chen2023neurbf, Li2022DCCDIF, airnets, martel2021acorn, mueller2022instantngp, chabra2020deeplocalshapes}.

We posit that positioning learnable elements in signal coordinate space can aid INR classification, as by coupling latent information to local image regions, we retain image structure in the latent space and the image features encoded therein. Other methods which position latents freely in image coordinate space focus on improving the signal reconstruction quality \citep{chen2023neurbf, airnets}. These methods hybridise INR methods \citep{chen2023neurbf}, and require intricate initialisation and latent decoding schemes \citep{chen2023neurbf, airnets}. In contrast, simplicity is a core design principle in \inr{}; reducing computational complexity ensures that fitting an entire image dataset remains efficient. 

\textbf{INR classification.} 
INR literature typically emphasises training and parameter efficiency whilst optimising reconstruction quality \citep{chen2023neurbf, mueller2022instantngp, dupont2021coin, huang2022compressingweather, essakine2024wherestandoninrs}, or utilises INRs as a component in a broader method for their compression and generalisation abilities \citep{cole2023andereapplic, dollinger2024sat}. Here, we investigate INRs in a classification setting.


Early works on INR classification studied flattened weight matrices \citep{unterthiner2020predicting} or their weights' statistics \citep{eilertsen2020ClassifyingTC}. Such flattened representations remain in use in INR-context as low-dimensional embeddings that are trained along with \citep{datatofuncta, spatialfuncta}, or after \citep{deluigi2023inr2vec}, INR fitting. These methods do not generalise well to image classification \citep{deluigi2023inr2vec, kalogeropoulos2024scaleequivariant} or larger-scale classification tasks \citep{spatialfuncta} however. Instead, a key insight to process \textit{whole} INR weight-spaces was to consider a neural network's symmetries; transformations which alter the weights but preserve the INR's function \citep{godfrey2022symmetries}. Architectures which incorporate equivariances to such symmetries obtain a significant increase in INR classification accuracy \citep{dwsnets, zhou2024neural, kofinas2024graph_nnsaregraphs, kalogeropoulos2024scaleequivariant}. With \inr{}, we propose an architecture that anchors low-dimensional latent embeddings in image-space. This ties the learnt encodings directly to local image content, effectively compressing an image into a latent point cloud. A similar observation is made and implemented in a concurrent work with an attention-based architecture \citep{wessels2024grounding}. Their method is demonstrated across various domains but does not address the shortcomings of INR classification concerning image classification, such as confinement to low-resolution image datasets and sensitivity to image-space translations.

\textbf{Flexibility of INRs.} The flexibility of NNs allows for wildly varying weight-spaces that faithfully capture a signal. This variability makes it difficult for downstream models to capture consistent image features across INR instances \citep{kofinas2024graph_nnsaregraphs, shamsian2024inr_augment, kalogeropoulos2024scaleequivariant}. Previous work has shown that establishing a form of alignment or `common ground' among INRs improves classification accuracy. Such methods include sharing the INR weight initialisation \citep{dwsnets, papa2024trainnef}, introducing shared learnable elements to the fitting process \citep{airnets, chen2023neurbf, wessels2024grounding}, learning low-dimensional shifts to an established INR base network \citep{datatofuncta, spatialfuncta} or sharing a part of the INR over all instances \citep{Vyas2024cutsirenintwo}. In this spirit, \inr{} shares a decoder over the whole dataset, which is jointly pretrained on a subset of the data and then frozen.

Even if INRs share a form of alignment, overfitting remains a persistent issue in INR classification \citep{kofinas2024graph_nnsaregraphs, shamsian2024inr_augment,  kalogeropoulos2024scaleequivariant}. A limited set of weight-space augmentations are available to combat this \citep{dwsnets, shamsian2024inr_augment}. Hence, INR classification methods fit redundant INRs for each image in the dataset, which is a resource-intensive task \citep{zhou2024permutation, kalogeropoulos2024scaleequivariant, dwsnets, spatialfuncta, groundingandenchance_2024}. Instead, we can leverage the unique weight-space of \inr{} to apply intuitive data augmentation methods on-the-fly which we demonstrate to be competitive in regularisation effectiveness to redundant INR fitting, at a fraction of the computational cost.


\section{Method: \inr{}}
\label{sec:method}


\inr{} consists of an encoder and a decoder. The encoder is mainly a cloud of latent vectors and retrieval logic to obtain the $n$ nearest latent vectors for a given input coordinate. These vectors are then concatenated and passed to the decoder, which maps the vector to a colour. More specifically, given a coordinate $\bm{x}\in \mathbb{R}^{d_{\text{in}}}$, an indexing function $U_n$ retrieves the $n$ nearest anchored latents and their relative positions to $\bm{x}$ from latent cloud $\mathcal{P}=\{ (\bm{p}_i, \bm{w}_i) \}_{i=1}^{k}$. These latents and their relative positions are concatenated. The concatenated vector is then passed through the MLP decoder $g$, mapping it to an RGB colour. Further details are in the Appendix. 

\begin{wrapfigure}{R}{0.45\textwidth}
    \centering
    \vspace{-1cm}
    \includegraphics[width=\linewidth]{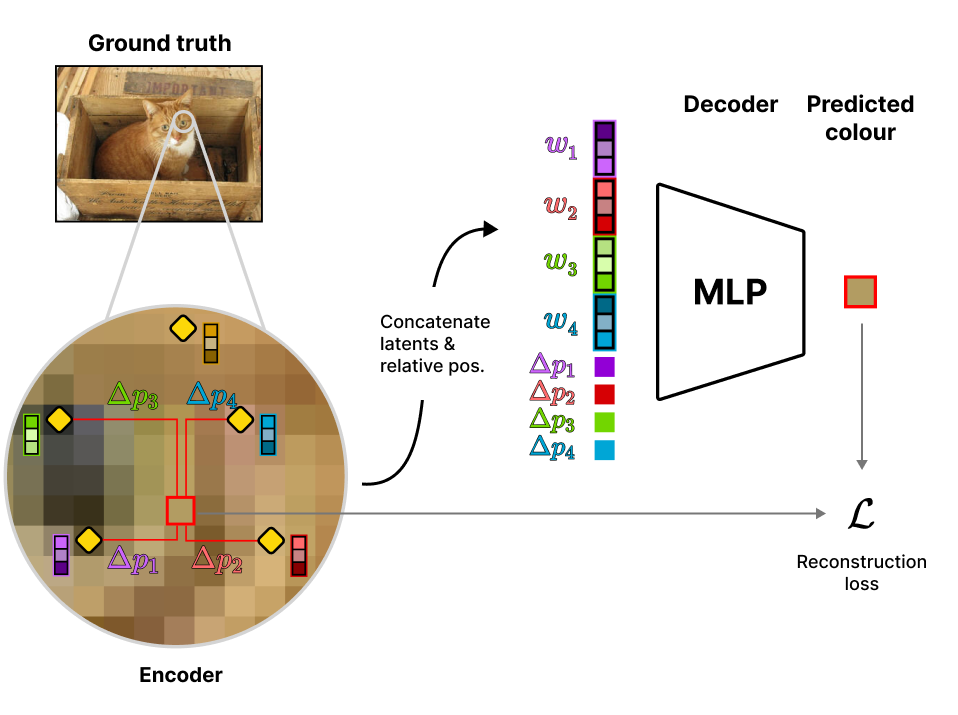}
    \caption{Given an image coordinate $\bm{x}$, \inr{} finds the 4 nearest latent vectors, along with their relative position to $\bm{x}$. These are concatenated into a long descriptive vector which the decoder maps to an RGB colour. The reconstruction loss is computed between the ground-truth and predicted colour.}
    \vspace{-1cm}
    \label{fig:method}
\end{wrapfigure}

\paragraph{Encoder.}
The encoder consists of a cloud of learnable latent vectors and aggregation logic. The latent vectors are anchored in the image coordinate space. The latent dimension $z$ and the number of latents $k$ anchored in the image are hyperparameters. 

\paragraph{Latent vector positions.} We follow \citet{chen2023neurbf, Li2022DCCDIF}, whereby learnable elements are positioned in signal-space near high-frequency content so as to bias the model capacity to more difficult to encode content. To this end, the latents' positions are determined by sampling the image gradient norm. The latents' positions remain fixed. The indexing function $U_n$ can therefore \textit{cache} the index of nearest latents upon \inr{} initialisation, significantly decreasing training latency.

\paragraph{Indexing and aggregation.} Given an input coordinate $\bm{x}$, an indexing function $U_n$ retrieves the $n$ nearest latent vectors along with their relative position to $\bm{x}$. $n$ can be any value but we found 4 nearest neighbours to be sufficiently expressive. In contrast to other methods which predefine an interpolation function to aggregate the latent vectors \citep{chen2023neurbf, airnets}, \inr{} defers to the decoder to learn this from the latent vectors and relative positions, similar to \citep{chen2021liif}. The latent vectors and their relative positions are thus simply concatenated and fed to the decoder. We found Fourier features \citep{tancik2020fourfeat, chen2023neurbf} to not improve results much so these were omitted in favour of simplicity.

\paragraph{Decoder.}
The decoder is a simple MLP with ReLU activations, as opposed to more elaborate activation functions \citep{chen2023neurbf}. To align latent vectors across different \inr{} instances, we let instances share the decoder. This shared decoder is pre-trained along with several \inr{}s on a subset of the data and then frozen for the complete dataset. Consequently, the memory cost of the decoder can be amortised across the whole dataset as only the anchored representation cloud is required for classification.

\paragraph{Downstream processing.}
Due to its unique weight-space, \inr{} transforms the problem of INR classification into point cloud classification. This allows us to leverage well-studied downstream architectures for \inr{} classification. Contrary to SIREN classifiers, no further mechanisms against weight-space equivariances are needed \citep{dwsnets, kofinas2024graph_nnsaregraphs, kalogeropoulos2024scaleequivariant}.

\textit{Point Transformer v3.} Any point cloud architecture that supports arbitrary point feature dimensions can naturally process \inr{}s. However, since the anchored latent vectors encode local information, an architecture that emphasises local interaction is preferred. To this end, we select Point Transformer v3 (PTv3) \citep{wu2024ptv3}, a state-of-the-art method that performs local attention. When using PTv3, we provide it only the learnt latent vectors. The latent positions are used only for the relative positional encoding and pooling operations. PTv3 allows for a varying number of points within a batch, enabling \inr{}s to adjust the number of latents to the image complexity. Note however that in this work we follow \citep{chen2023neurbf} by letting the number of latent vectors be proportional to image size. Modifications made to PTv3 for \inr{} compatibility are discussed in \cref{sec:app_inr_implementation}.

\paragraph{Data augmentation.} We can leverage the unique weight space of \inr{} to apply intuitive data augmentation methods which we show to be almost as effective as using redundant \inr{} instances in our experiments. These data augmentations are applied `on-the-fly' on the anchored representation cloud. The augmentation methods are named Noise, Mask, Push, Rotate, Flip, and \inr{}mix (\autoref{fig:data_augmentation}). Noise applies random Gaussian noise to the latent content, Masking omits a specified fraction of the points, Push, Rotate and Flip only augment the latent vector coordinates within a single \inr{} instance. Furthermore, \inr{}mix, a method inspired by CutMix \citep{yun2019cutmix, zhang2022pointcutmix}, mixes two \inr{} by combining their latent clouds.


\section{Experiments}
\label{sec:experiments}

In INR classification literature, the seminal SIREN architecture remains the most prominent and has seen incremental classification improvements over recent years \citep{dwsnets, kofinas2024graph_nnsaregraphs, zhou2024permutation, zhou2024neural, kalogeropoulos2024scaleequivariant}. We focus on two representative baselines: the foundational DWSnets \citep{dwsnets} and the state-of-the-art ScaleGMN \citep{kalogeropoulos2024scaleequivariant}. Additional experiments are presented in \cref{sec:app_further_experiments}, along with more details about the experiments and implementation.

\paragraph{Experiment 1. High-resolution image classification.}
How well do \inr{}s and existing INR classification pipelines perform as image resolution increases? To answer this question, we analyse INR classification accuracy on Fashion-MNIST (FMNIST) \citep{fashionmnist} whereby we pad the images with zeroes to a $100 \times 100$ and $1024 \times 1024$ resolution (\autoref{fig:app_highres_preview}). These datasets are then converted into SIRENs and \inr{}s and classified by their respective methods. The test accuracy is reported in \autoref{tab:exp_highres_fmnistclass}. 

While both SIREN and \inr{} classification pipelines show a degradation in classification accuracy, \inr{} is demonstrably stronger. This difference can be attributed to how each method handles increasing image resolution. A SIREN's input domain is fixed to $[-1, 1]^{d_{\text{in}}}$, regardless of image size. On higher resolutions, the number of pixels mapped within $[-1, 1]^{d_{\text{in}}}$ grows, requiring SIRENs to learn higher frequency mappings. Subsequently, spectral bias effects re-emerge, resulting in overly smooth reconstructions. To mitigate this, the SIREN architecture is increased in width and depth, yielding a $\times33$ increase in the parameters between $100 \times 100$ and $1024 \times 1024$. This substantial increase significantly increased fitting time, requiring us to limit the SIREN dataset size to approximately a third of that used for \inr{}. For \inr{}, the image resolution is decoupled from the latent representation. Consequently, as image size grows but image complexity remains low, the number of latent vectors does not have to be adapted.


\begin{table}[b]
    \centering
    \small
    
    \begin{tabular}{lcccc}
\toprule
    \textbf{Side Length} & \textbf{28} & \textbf{100}  & \textbf{1024} & \textbf{\makecell{INR \#param\\increase}}\\
\midrule
    DWSNets     &   67.06$^\diamond$            & 67.60         & 53.28         & $\times$33  \\
    ScaleGMN    &   \textbf{80.78}$^\diamond$   & 74.50         & 48.77         & $\times$33  \\
    Ours        &   80.42                       & \textbf{79.36}& \textbf{73.57}& \textbf{$\times$1} \\
\bottomrule
    \end{tabular}
    \caption{\textbf{Exp. 1:} Test accuracy (\%$\uparrow$) on the padded FMNIST datasets and the required increase in INR parameters ($\downarrow$) to produce recognisable reconstructions. The SIREN $1024 \times 1024$ dataset is a third of the size of the corresponding \inr{} dataset due to steep fitting costs as a consequence of the increased number of parameters. Entries marked with $^\diamond$ are taken from their original publication \citep{dwsnets, kalogeropoulos2024scaleequivariant}. Next to being a more parameter efficient representation, our method is more resilient against increasing image size.}
    \label{tab:exp_highres_fmnistclass}
\end{table}

We continue our investigation of high-resolution image INR classification with Imagenette \citep{imagenette}, a subset of Imagenet \citep{krizhevsky2012imagenet}. Imagenette is a dataset of natural images with a median resolution of $375 \times 500$ and a maximum of $4268 \times 2912$. As image complexity increases, INR capacity must increase proportionally. For \inr{}, we can increase the number of anchored latent vectors. For SIRENs however, the scaling is performed in either the number of hidden layers or the hidden dimension. Neither DWSnets nor ScaleGMN support variable SIREN architectures, so a fixed architecture size must be picked. This invariably leads to undercapacity or overcapacity on a subset of the image data. To decrease discrepancies that may arise from this, we use the prescaled dataset variant Imagenette320 \citep{imagenette}, and apply a centre-crop to standardise all images to $320 \times 320$. These images are converted into SIRENs and \inr{}s and subsequently classified. This procedure is additionally performed on the original, full-resolution Imagenette dataset with \inr{}s. As no test set is provided, we report validation accuracy in \autoref{tab:exp_imagenette_acc}. We were not able to train ScaleGMN on this dataset due to instability issues, which we discuss further in \cref{sec:app_expdetails}.

\inr{} demonstrates high classification accuracy on Imagenette. We hypothesise that this performance is due to the higher resolution of Imagenette images, where relevant features are distributed across larger regions of the image. With latent vectors anchored in these regions, \inr{} can represent the features with greater precision and redundancy, all while respecting the spatial integrity of these features. This in turn enables PTv3 to infer meaningful image features. The additional accuracy between the centre-crop and full-resolution sets can be attributed to the resolution bias present in Imagenette, which may be exploited by the relative positional encoding and pooling in PTv3; in a brief test we found that a simple ReLU-MLP of dimensions $[2, 64, 64, 64, 10]$ can obtain a validation accuracy of up to $23.46 \%$ on Imagenette when trained on image dimensions alone.

\begin{table}
    \small
    \centering
    \begin{tabular}{lcc}
\toprule
                        & \textbf{\makecell{Imagenette 320x\\CenterCrop}} & \textbf{\makecell{Imagenette full\\resolution}} \\
\midrule
    DWSnets             & 41.05                         & - \\
    Ours                & \textbf{71.71}                & \textbf{75.92} \\
\bottomrule
    \end{tabular}
    \caption{\textbf{Exp. 1:} Validation accuracy (\%$\uparrow$) on Imagenette. \inr{} sets a new watermark in classifying full-resolution image data through their INR representation.}
    \label{tab:exp_imagenette_acc}
\end{table}

\paragraph{Experiment 2. Image classification benchmarks.}
How does \inr{} compare to established INR classification benchmarks? We follow other INR literature in using MNIST \citep{lecun1998mnist}, Fashion-MNIST \citep{fashionmnist}, and CIFAR10 \citep{krizhevsky2009cifar}. The complete datasets are used, along with data augmentation methods that are expanded upon in Experiment 3. Results are gathered on 3 different seeds and summarised in \autoref{tab:classification}. On low-complexity grey-scale datasets, \inr{} performs similarly to current state-of-the-art methods. On the more complex CIFAR10 dataset, \inr{} obtains state-of-the-art accuracy. CIFAR10 contains natural RGB images which inherently contain background noise. It is therefore a more challenging benchmark than the gray-scale MNIST and FMNIST datasets. We expect that image features are better represented among \inr{} instances on these more complex images, leading to superior accuracy compared to SIRENs.

\begin{table}
    \small
    \centering
    \begin{tabular}{lcccc}
\toprule
     &  \textbf{MNIST} & \textbf{FMNIST}   &  \textbf{CIFAR10}           \\
\midrule
    DWSnets  \citep{dwsnets}   &  85.71 $\pm$ 0.6   &  67.06 $\pm$ 0.3  &  -                           \\
    NG-GNN   \citep{kofinas2024graph_nnsaregraphs}   &  91.40 $\pm$ 0.6    &  68.00 $\pm$ 0.2  &  36.04$^\diamond$ $\pm$ 0.44 \\
    NG-T     \citep{kofinas2024graph_nnsaregraphs}   &  92.40 $\pm$ 0.3    &  72.70 $\pm$ 0.6  &  - \\
    ScaleGMN \citep{kalogeropoulos2024scaleequivariant}   &  \textbf{96.59} $\pm$ 0.2   &  \textbf{80.78} $\pm$ 0.2  &  38.82 $\pm$ 0.1            \\
    
    Ours        &  92.69 $\pm$ 1.2  &  \textbf{80.42} $\pm$ 0.4  & \textbf{58.47 $\pm$ 0.4}  \\
\bottomrule
    \end{tabular}
    \caption{\textbf{Exp. 2:} Test classification accuracy (\%$\uparrow$) on various image classification datasets. Entries marked with $^\diamond$ are taken from reproductions by \citep{kalogeropoulos2024scaleequivariant}. \inr{} classification accuracy is similar to baselines on low-complexity datasets and outperforms them on the more complex CIFAR10 dataset.}
    \label{tab:classification} 
\end{table}

\paragraph{Experiment 3. Data augmentation.}
In SIREN classification methods, an increasingly used technique to reduce overfitting is to generate redundant INRs for each image in the dataset \citep{zhou2024permutation, kalogeropoulos2024scaleequivariant, dwsnets, spatialfuncta, groundingandenchance_2024}, demanding significantly more resources and time for the INR fitting process. To establish a baseline, we fit 20 \inr{}s per image on a 10k subset of CIFAR10. Without any additional augmentations, PTv3 is trained on two conditions: using a single \inr{} per image and using all 20 \inr{}s. We report the test accuracy per image in \autoref{tab:augment_redundancy}. \inr{} classification accuracy is on par with the state-of-the-art which has the advantage of being trained on the full CIFAR10 dataset.

We now ask, are data augmentation methods that operate on the \inr{} weight-space as effective as using redundant INRs during training? We convert the entire CIFAR10 dataset to \inr{} instances and evaluate the different data augmentation methods in \autoref{tab:augment_weightspace}. When comparing the test accuracies obtained under weight-space data augmentations to those obtained with redundant INR fitting \autoref{tab:augment_redundancy}, we observe that our data augmentations do not fully close the gap. Moreover, the gap may be slightly larger as we leverage the entire CIFAR10 in the weight-space augmentation experiments and just a 10k subset in the redundant INR experiment. Regardless, our data augmentation methods offer a compelling alternative that requires significantly less computational resources and time to execute.

\begin{table}
    \begin{minipage}[b]{.45\linewidth}
        \small
        \centering
        \begin{tabular}{lcc}
            \toprule
                \textbf{\#INRs per image} & \textbf{1} & \textbf{20} \\
            \midrule
                NG-GNN      & 36.04$^\diamond$ & 45.70$^\diamond$ \\
                ScaleGMN    & \textbf{38.82} & \textbf{56.95} \\
                Ours        & 38.12 & 55.87  \\
            \bottomrule
        \end{tabular}
        \caption{\textbf{Exp. 3:} CIFAR10 test accuracy (\%$\uparrow$) after training on either 1 or 20 INRs per CIFAR10 image. No further data augmentation is employed. Similar to the baselines, \inr{} classification accuracy improves significantly when trained on redundant INRs. Baseline data is taken from \citet{kalogeropoulos2024scaleequivariant}, where data marked with $^\diamond$ denote their reproduction of \citep{kofinas2024graph_nnsaregraphs}. }
        \label{tab:augment_redundancy}
    \end{minipage} \hfill
    \begin{minipage}[b]{.52\linewidth}
    \small
    \centering
    \begin{tabular}{cccc|c}
    \toprule
        \textbf{\makecell{Latent\\noise}} & \textbf{\makecell{Point\\space}} & \textbf{\makecell{\inr{}\\masking}} & \textbf{\inr{}mix} & \textbf{Acc.} \\ 
    \midrule
        -           &  -            & -         &  -            & 38.12 \\
        \checkmark  &  -            & -         &  -            & 39.08 \\
        -           & \checkmark    & -         &  -            & 51.69 \\
        -           &  -            &\checkmark &  -            & 50.25 \\
        -           &  -            & -         & \checkmark    & 54.55 \\
        -           &  -            &\checkmark & \checkmark    & 54.56 \\
        -           & \checkmark    &\checkmark & \checkmark    & 50.34 \\
        \checkmark  & \checkmark    &\checkmark & \checkmark    & 51.03 \\
    \bottomrule
    \end{tabular}
    \caption{\textbf{Exp. 3:} Test accuracy (\%$\uparrow$) on different \inr{} augmentation methods. The `Point space' column applies the push, flip and rotation augmentation methods. Compared to fitting and training on redundant INRs, these more efficient weight-space augmentations yield competitive test accuracy.}
    \label{tab:augment_weightspace}
    \end{minipage} 
\end{table}

\begin{figure}
    \centering
    \includegraphics[width=0.97\linewidth]{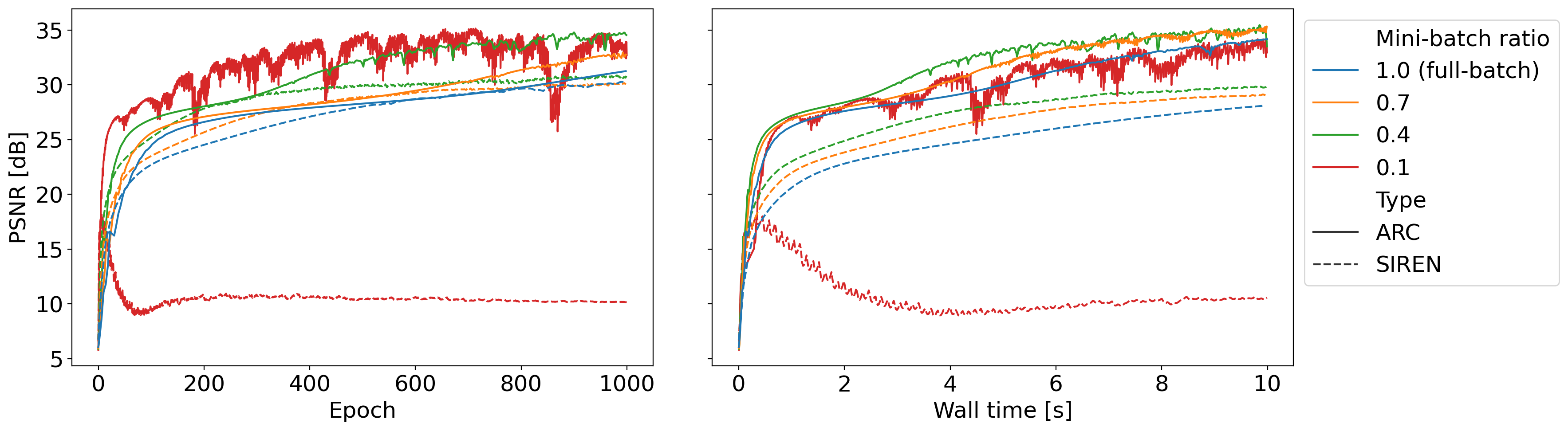}
    \caption{\textbf{Exp. 4:} 
    Image reconstruction quality when trained with various mini-batch ratios. Interestingly, both SIREN and \inr{} show strong fitting performance when only using a subset of the pixels at each optimization step. This allows for achieving a higher PSNR in less wall time.}
    
    \label{fig:app_mini-batches}
\end{figure}

\vspace{1cm}
\paragraph{Experiment 4. Mini-batch training.} \label{sec:exp_mini-batches}
In INR context, training for an epoch entails having supervised all image coordinates. We ask whether mini-batches can be an alternative, whereby optimisation steps are performed on a random subset of the image coordinates. To answer this question, we define \textit{mini-batch ratio} as the fraction of the full-batch size used in a single step. For several mini-batch ratios, we train a SIREN and \inr{} instance three times on skimage's astronaut image and average them together. In \autoref{fig:app_mini-batches} we observe that, for both SIRENs and \inr{}s, mini-batch training is stable for high enough ratios, and yields similar reconstruction quality as full-batch training in fewer epochs and wall time. Encouraged by this finding, we leverage mini-batch training in all our \inr{} experiments with a mini-batch ratio of 0.25 unless otherwise noted.

\paragraph{Experiment 5: Image translation robustness.}
For regular image classifiers, translation invariance means that the classifier's prediction is unaffected by a shift in pixels. This runs counter to INRs where any change in pixels should be accurately reflected in the reconstruction, and consequently, in the learnt INR weights. For an INR classifier to be robust against benign image shifts, it must distinguish changes in INR weights that capture pixel shifts from changes that capture more meaningful semantic changes.

\newcommand{\centeredZero}{\textit{\monospace{Centered}}}
\newcommand{\displacedZero}{\textit{\monospace{Displaced}}}

\begin{table}
    \centering
    \small
    \footnotesize
    \begin{tabular}{lcc}
    \toprule
        \textbf{Method} & \textbf{\makecell{Test on\\\centeredZero{}}} & \textbf{\makecell{Test on\\\displacedZero{}}} \\ 
    \midrule
        ScaleGMN    &       74.50       & 13.00 \\ 
        Ours, PTv3  &       79.36       & 47.61 \\
    \bottomrule
        \end{tabular}
    \captionsetup{type=table}
    \caption{\textbf{Exp. 5:} Test accuracy (\%$\uparrow$) when trained on INRs of \centeredZero{} and evaluated on INRs of \displacedZero{}. SIREN-based methods (ScaleGMN) experience a complete collapse in classification accuracy due to significant differences in weight-space among the datasets. \inr{} paired with PTv3 demonstrates improved robustness but warrants further experimentation.}
    \label{tab:transinvar_0s}
\end{table}

As \inr{} captures local image features, we hypothesise that the weight-space remains largely intact when fitting to a shifted image, making the \inr{} classification pipeline more robust to such transformations. To test this, we return to the $100 \times 100$ padded Fashion-MNIST objects, where, alongside centred FMNIST objects (named \centeredZero{}), we fit INRs to randomly displaced FMNIST images (named \displacedZero{}). A sample of these data can be found in \autoref{fig:transinvar_0s}. SIRENs share the same initialisation, while \inr{}s use the same decoder over all datasets. Both ScaleGMN and PTv3 are trained on their respective INRs of the \centeredZero{} dataset and subsequently evaluated on \displacedZero{} INRs. Test accuracies are listed in \autoref{tab:transinvar_0s}.


In SIREN-based methods, evaluating on \displacedZero{} is almost equivalent to random guessing, as the displaced FMNIST images have induced a drastic change to the SIREN weight-space. For \inr{}, we observe a much smaller, but still significant drop in generalising to \displacedZero{} \inr{}s. Like with SIRENs, the shifted image content may have induced significant differences in the latent content. However, an additional cause may be at play, where PTv3 is overfitting to the absolute positions of the \inr{} latent vectors. Although PTv3 does not explicitly use absolute latent positions, its pooling and local attention mechanisms \textit{do} rely on them. We test the trained PTv3 model on \centeredZero{} \inr{}s where the entire latent cloud is shifted. In \autoref{tab:abs_pos_overfitting}, this shift causes a significant drop in accuracy degrades, confirming that PTv3 is not translation invariant. We therefore also consider Point Transformer v1 (PTv1) which uses a simpler nearest neighbour mechanism relying on relative position \citep{zhao2021pointtransformer}. PTv1 shows no significant changes in accuracy drop under the same conditions as PTv3.

To address the issue of translation sensitivity, we retrain PTv3 but randomly shift the \inr{}s during training. This forces PTv3 to become more robust to absolute position differences. When evaluated on \displacedZero{}, the accuracy drop is significantly reduced compared to the original setup, as shown in \autoref{tab:trained_with_shifts}. With the increased robustness against absolute position differences, the observed performance gap can be attributed to latent content differences between \centeredZero{} and \displacedZero{}. PTv1 is also tested but shows significantly weaker generalisation properties. 

\begin{table}
    \begin{minipage}[b]{.47\linewidth}
        \centering
        \small
        \begin{tabular}{lcc}
        \toprule
                        & \textbf{No shift} &  \textbf{Shift}   \\
        \midrule
             PTv3       & 79.32             & 57.87 \\
             PTv1       & 76.37             & 76.46 \\
        \bottomrule
        \end{tabular}
        \caption{\textbf{Exp. 5:} Impact of absolute latent position shifts on PTv3 and PTv1 test accuracy ($\uparrow$\%). PTv3 shows a substantial drop in accuracy when the latent cloud is shifted, demonstrating its sensitivity to absolute latent positions. PTv1 shows no noticeable accuracy degradation which is in line with its relative position mechanisms.}
        \label{tab:abs_pos_overfitting}
    \end{minipage} \hfill
    \begin{minipage}[b]{.47\linewidth}
        \centering
        \small
        \begin{tabular}{lcc}
        \toprule
            \textbf{Method} & \textbf{\makecell{Test on\\\centeredZero{}}} & \textbf{\makecell{Test on\\\displacedZero{}}} \\ 
        \midrule
             PTv3       & 78.58       & 62.78 \\
             PTv1       & 76.37       & 49.36 \\
        \bottomrule
        \end{tabular}
        \caption{\textbf{Exp. 5:} Test accuracy ($\uparrow$\%) when training PTv3 on \centeredZero{} with random latent cloud shifts. Naturally, training with latent cloud shifts improves PTv3’s robustness to translation and allows us to analyse the accuracy gap due to latent content differences between \centeredZero{} and \displacedZero{}. PTv1 has this property built in but shows significantly weaker generalisation across the datasets.}
        \label{tab:trained_with_shifts}
    \end{minipage}
\end{table}

\section{Conclusion}
\label{sec:conclusion}

In this paper, we introduced \inr{}: Anchored Representation Clouds, a novel type of INR that retains image structure in its weight-space. \inr{} can leverage point cloud classification architectures to obtain state-of-the-art classification results and process image datasets that were previously unattainable for INR-based classification methods. Additionally, the unique weight-space of \inr{} provides efficient data augmentation techniques. For future work, making \inr{} adapt its capacity dynamically to the image's complexity would be an interesting avenue. Additionally, a key advancement in INR classification would entail the unification of the INR fitting and classification processes, which are currently treated as distinct stages. End-to-end coupling could lead to more competitive classification performance compared to image-space classification whilst being more memory-efficient.

\clearpage
\bibliography{iclr2025_conference}
\bibliographystyle{iclr2025_conference}

\clearpage
\appendix
\section{Appendix}

\subsection{ARC definition} \label{sec:app_arcdefinition}
We interpret a sampled signal $s$ as a set of equidistant discrete observations $\{ (\bm{x}_i \in \mathbb{N}^{d_{\text{in}}} , s(\bm{x}_i) \in \mathbb{N}^{d_{\text{out}}}) \}_{i=1}^{b}$. For instance, an RGB image would be a set of pixel locations ($d_{\text{in}}=2$) with corresponding RGB colours ($d_{\text{out}}=3$). An INR learns parameters $\theta$ by supervising the mapping between the signal's domain and codomain $f_{\theta}: \mathbb{R}^{d_{\text{in}}} \mapsto \mathbb{R}^{d_{\text{out}}}$, supervising on $s(\bm{x})$, with mean squared error (MSE) loss. \inr{} is defined as follows.

\footnotesize
\noindent
\begin{minipage}[c]{0.49\linewidth}
        
    \begin{flalign}
        f_\theta(\bm{x})= &g_\psi(e (\bm{x})) & \text{\scriptsize \inr{}} \label{eq:inr_def} \\
                          & \mathbb{R}^{d_{\text{in}}} \mapsto \mathbb{R}^{d_{\text{out}}} \\
        e(\bm{x})= &\text{Concat}(U_n(\bm{x}))  & \text{\scriptsize Encoder} \label{eq:encoder_def} \\
                   & (\mathbb{R}^z \times \mathbb{R}^{d_{\text{in}}})^n \mapsto \mathbb{R}^{n \cdot (z + d_{\text{\text{in}}})} \\
        g_{\psi}(\bm{v}_{\bm{x}})= &\text{MLP}_{\psi}(\bm{v}_{\bm{x}}) & \llap{\scriptsize Decoder} \label{eq:decoder_def} \\
                   & \mathbb{R}^{n \cdot (z + d_{\text{\text{in}}})} \mapsto \mathbb{R}^{d_{\text{out}}}
    \end{flalign}
\end{minipage} \hfill
\begin{minipage}[c]{0.49\linewidth}
    \begin{flalign}
        U_n(\bm{x})= &\{ (\Delta \bm{p}_i, \bm{w}_i) \}_{i=1}^{n} & \llap{\scriptsize Indexing function} \label{eq:indexing_def} \\
                 &\mathbb{R}^{d_{\text{in}}} \mapsto (\mathbb{R}^{d_{\text{in}}} \times \mathbb{R}^z)^n \\
    \mathcal{P}= & \{ (\bm{p}_i, \bm{w}_i) \}_{i=1}^{k}           & \llap{\scriptsize Latent cloud} \label{eq:latents_def} \\
                & \text{where } \bm{p}_i \in \mathbb{R}^{d_{\text{in}}}, \, \bm{w}_i \in \mathbb{R}^{z} \\
     \theta = &\{ \psi, \{\bm{w}_i\}_{i=1}^k \} & \llap{\scriptsize Learnable parameters} \text{ } \label{eq:learnables_def}
    \end{flalign}
\end{minipage}
\normalsize

\subsection{Further experiments.} \label{sec:app_further_experiments}
\paragraph{\inr{} feature locality.}
In this experiment, we test our claim that the anchored latent vectors of \inr{} represent local image features. If so, applying a transformation to the \inr{} coordinates should yield a similarly transformed reconstruction. Furthermore, in classifying \inr{}s, the latent vector positions should prove to be relevant. To remind the reader, the absolute position of the points is not given as a feature to PTv3 to learn from.

\begin{wraptable}{R}{0.35\linewidth}
    \centering
    \small
    \begin{tabular}{llcc}
    \toprule
           &  & \multicolumn{2}{c}{\textbf{Train}} \\
    \cmidrule{3-4}
           &        & Intact        & Pushed   \\
    \midrule
        \multirow{2}{*}{\rotatebox[origin=c]{90}{\makebox[0pt]{\textbf{Test}}}} 
           & Intact         & 79.32     & 51.79   \\
           & Pushed         & 17.39     & 69.06   \\
   \bottomrule
    \end{tabular}
   \caption{\textbf{Feature locality exp.:} FMNIST \inr{} test accuracy (\%$\uparrow$) if the latent vectors are displaced at either train or test time. Keeping the latent vectors' positions intact yields the highest accuracy, demonstrating their significance.}
    \label{tab:exp_coordinateshifts}
\end{wraptable}

\begin{figure}
\centering
\addtolength{\tabcolsep}{-4pt}    

    \begin{tabular}{cccc@{\hskip 15pt}cc}   
        Original & Shift & Rotate & Flip & Mask & Stack \\
        \includegraphics[width=2cm]{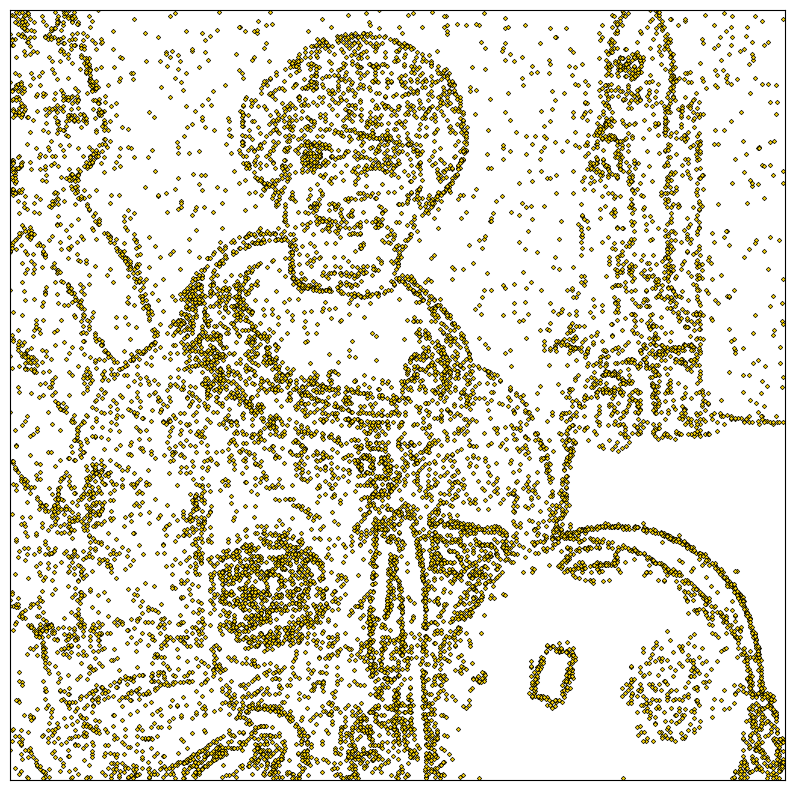} & \includegraphics[width=2cm]{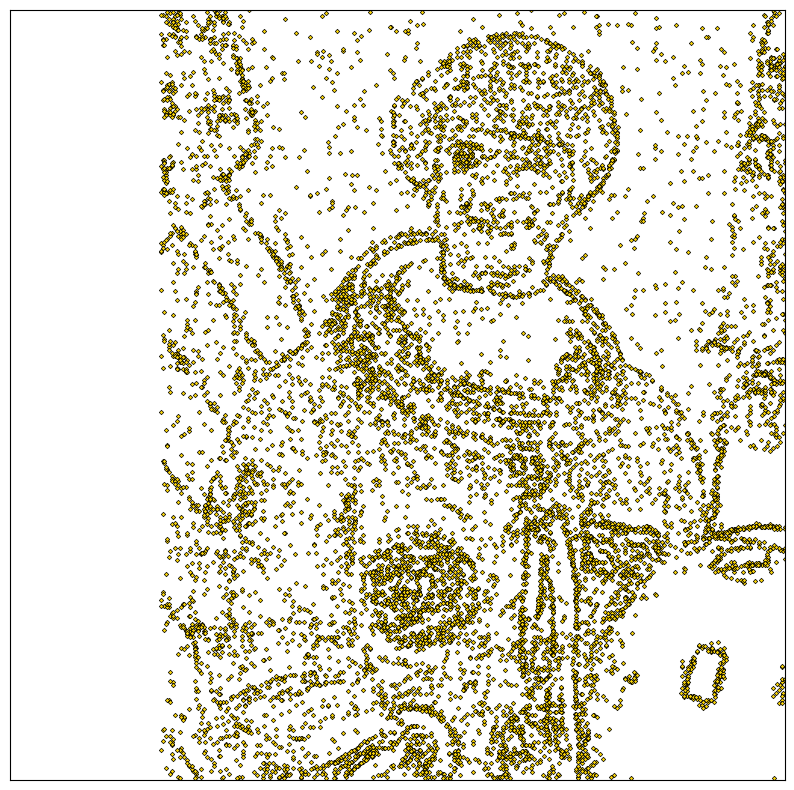} & \includegraphics[width=2cm]{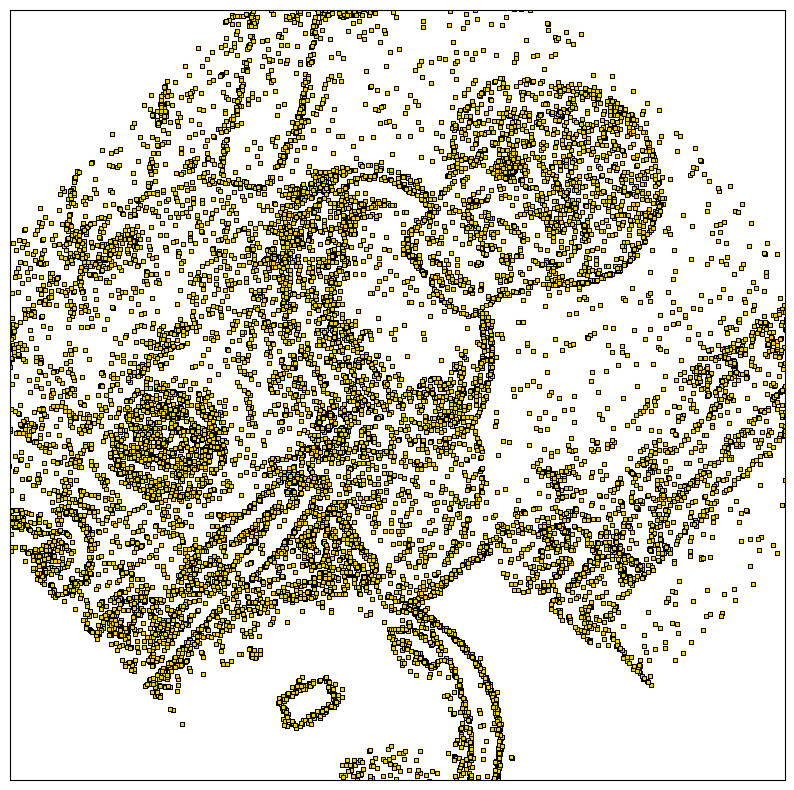} & \includegraphics[width=2cm]{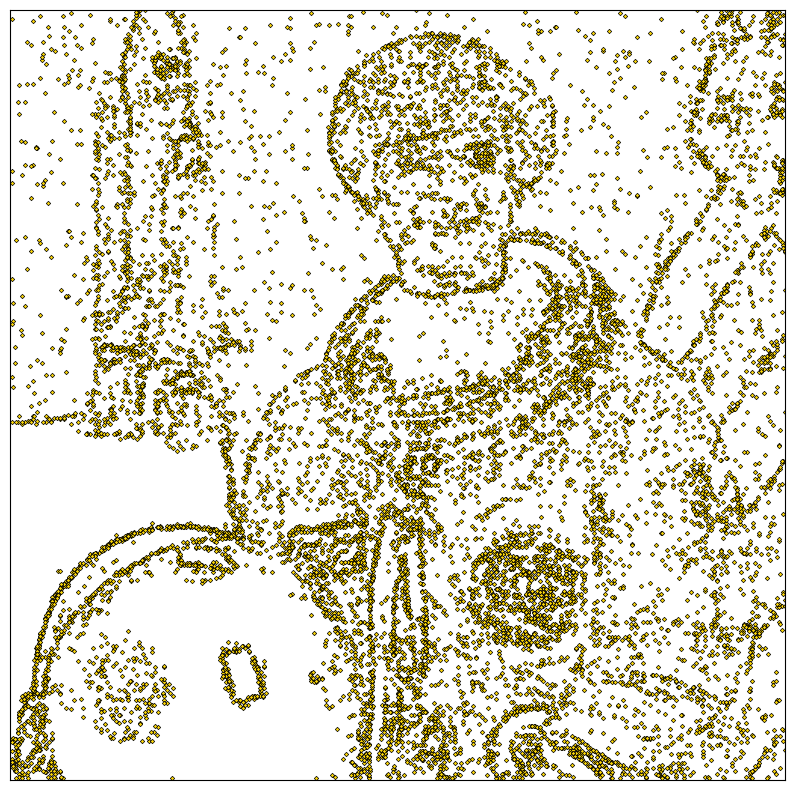} & \includegraphics[width=2cm]{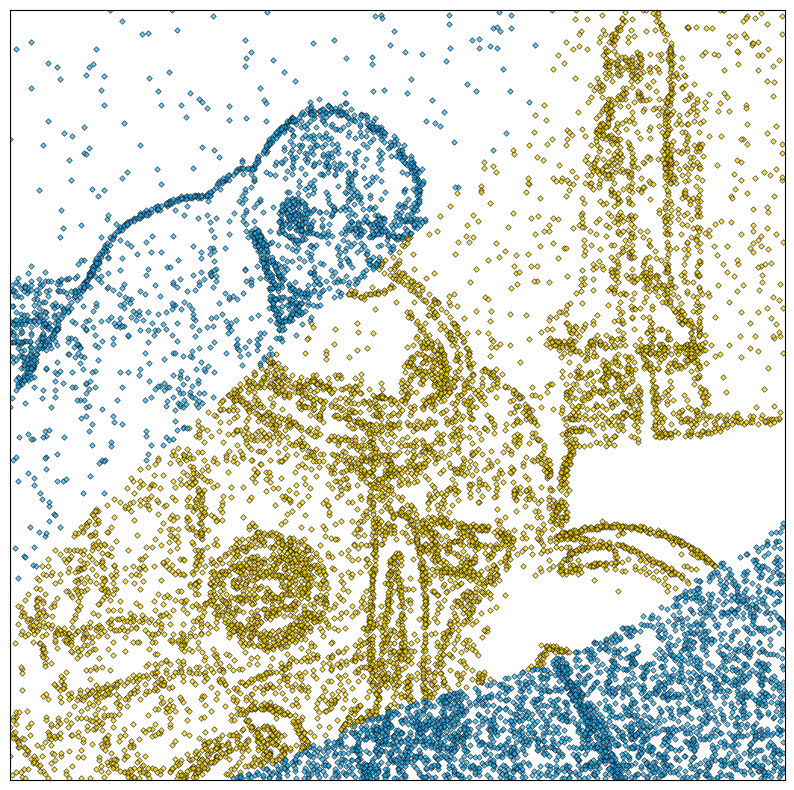} & \includegraphics[width=2cm]{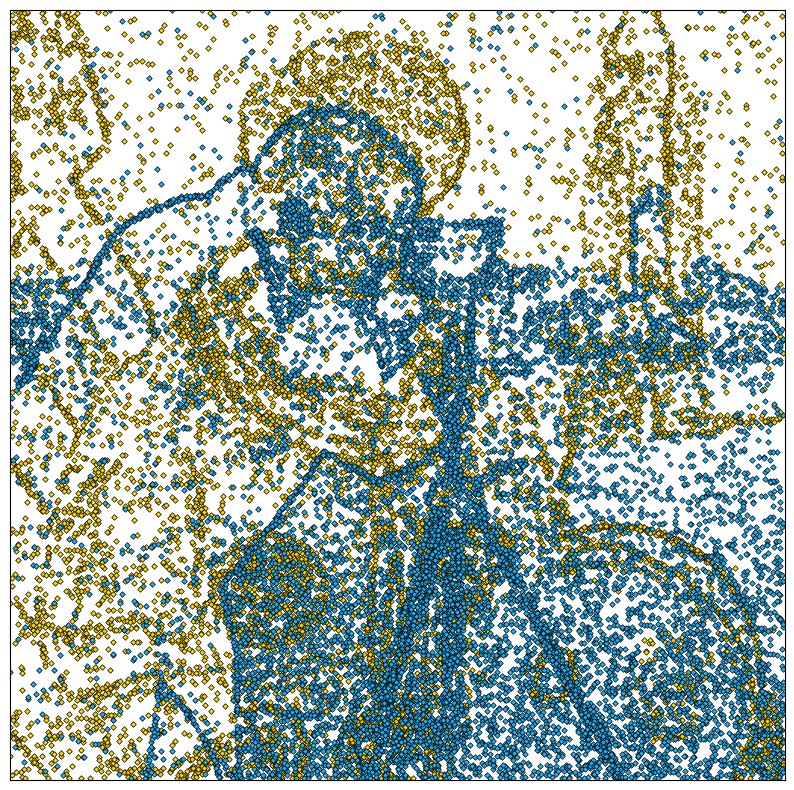} \\
        \includegraphics[width=2cm]{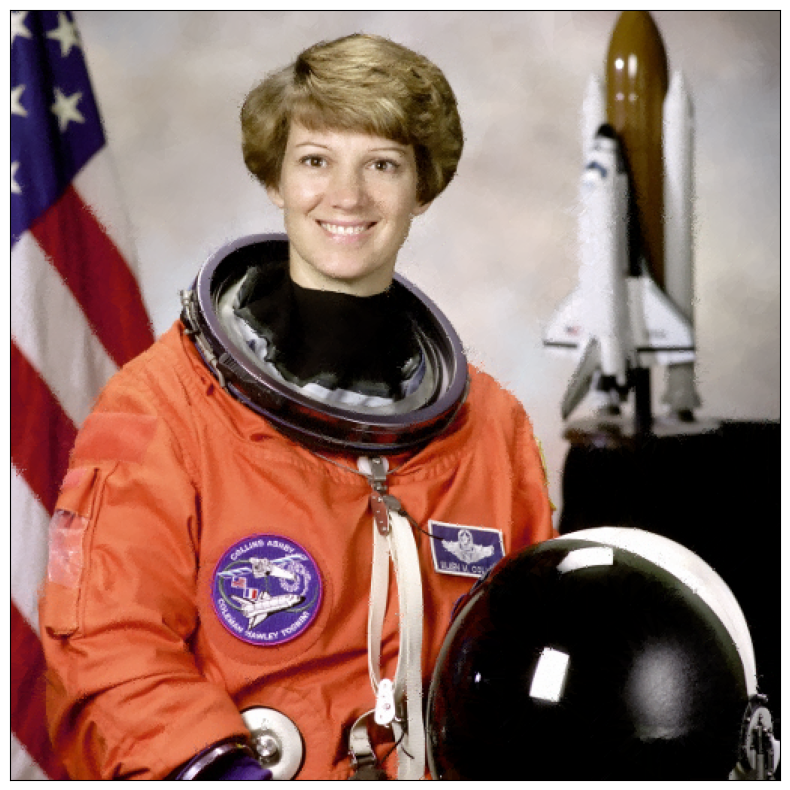} & \includegraphics[width=2cm]{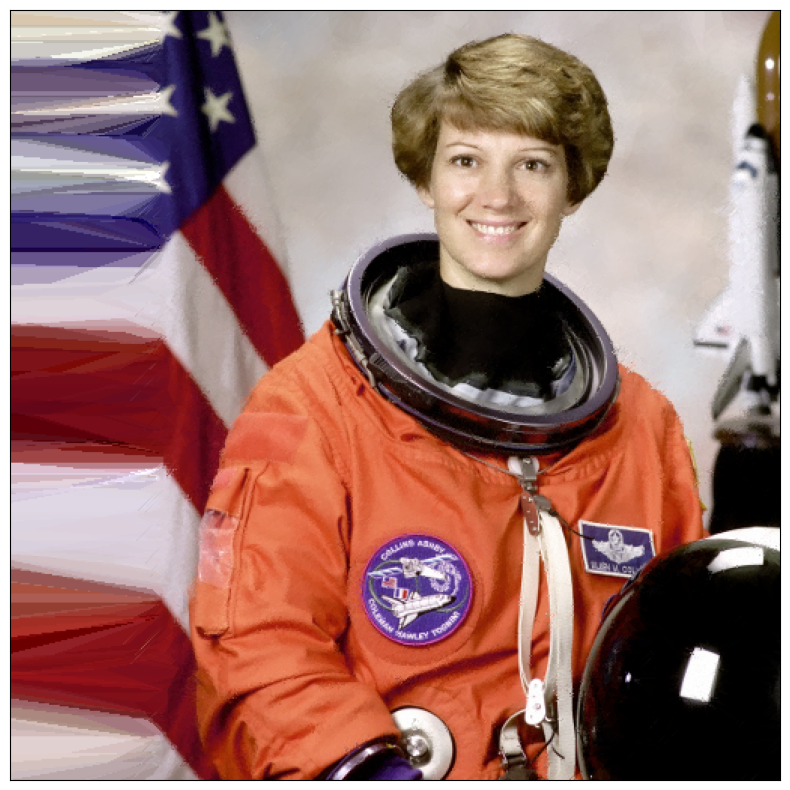} & \includegraphics[width=2cm]{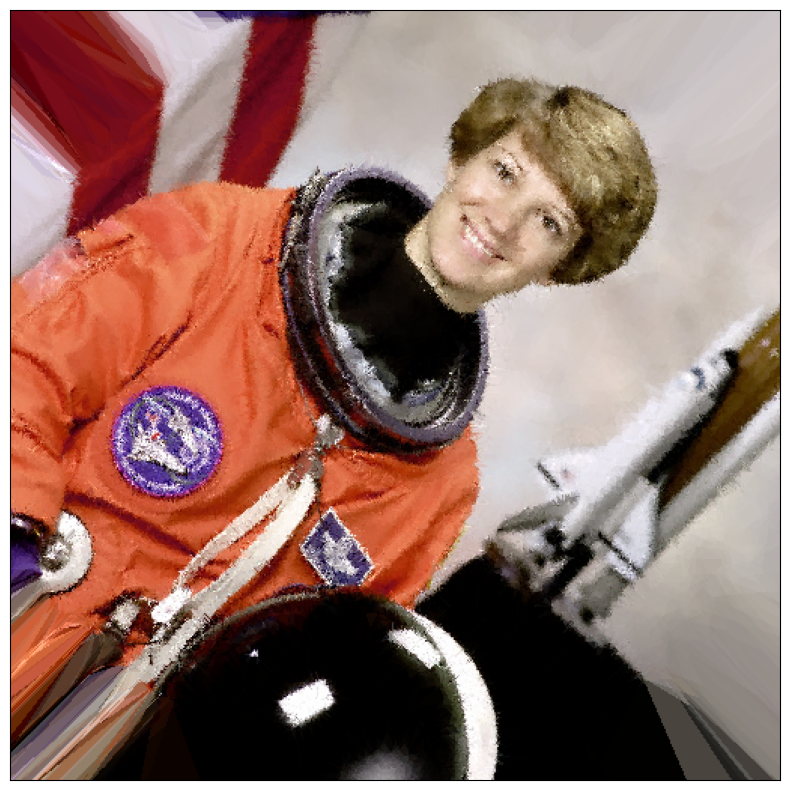} & \includegraphics[width=2cm]{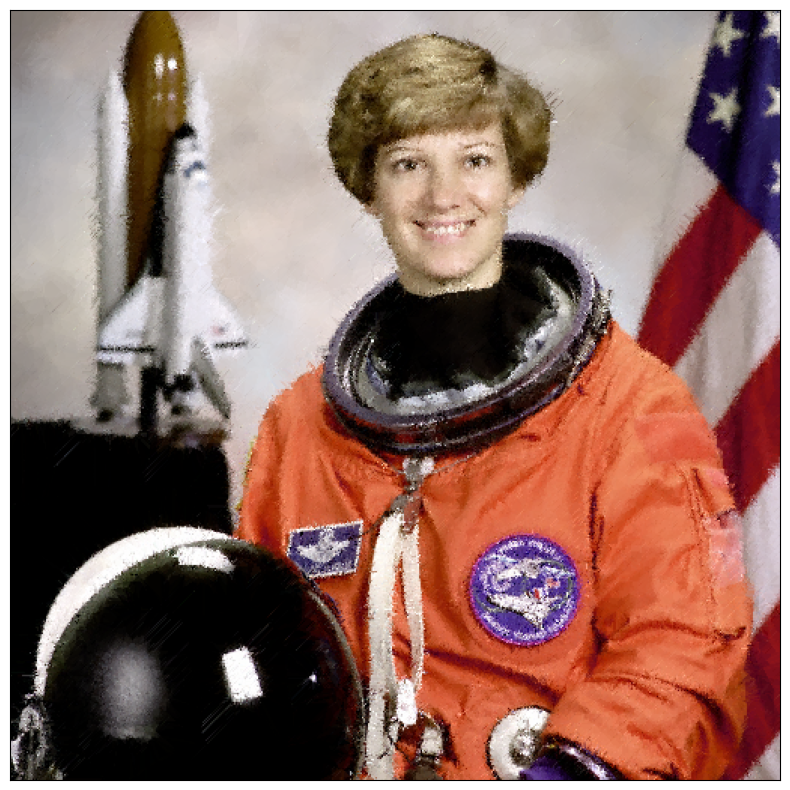} &  \includegraphics[width=2cm]{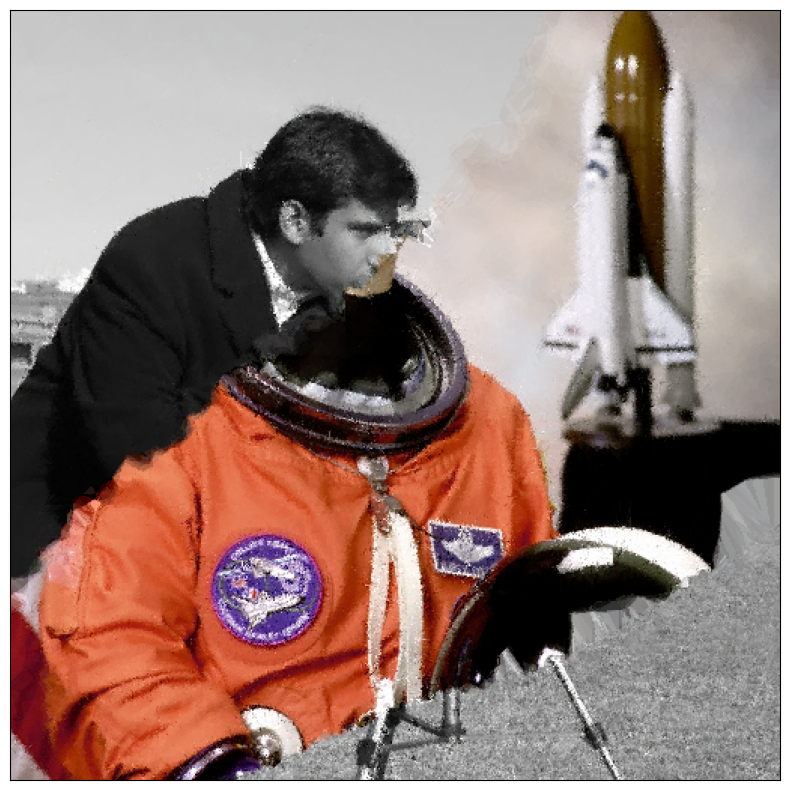} & \includegraphics[width=2cm]{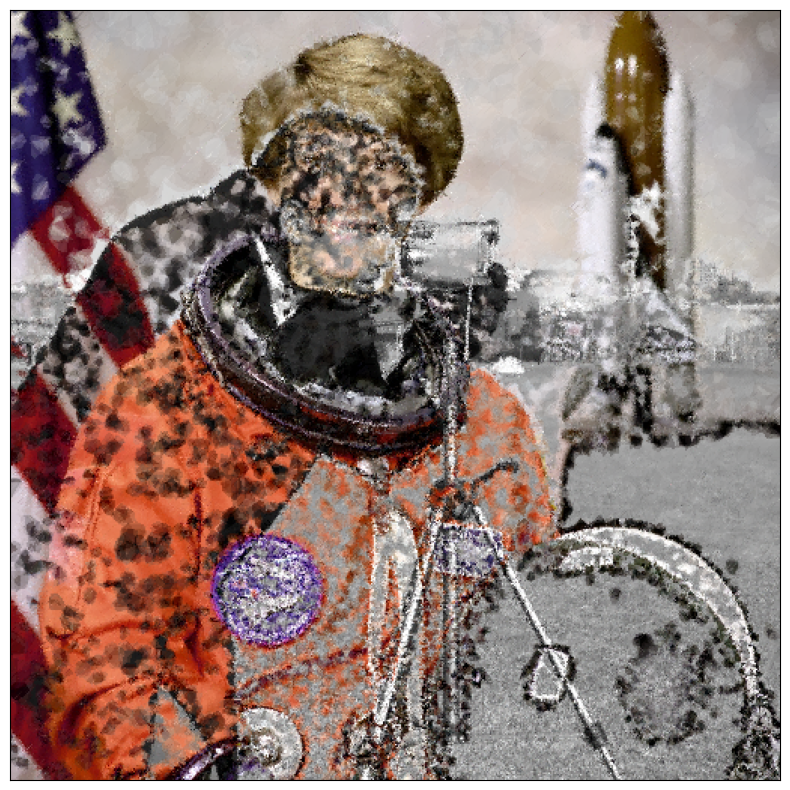} \\
       
    \end{tabular}
\addtolength{\tabcolsep}{4pt}   	
    \caption{\textbf{Feature locality exp.:} Each column depicts a transformation on the latent vector positions, where the bottom row shows the resulting construction. No other changes are made to the \inr{}. The correspondence between the latent cloud transformation and the new reconstructions demonstrates how \inr{} encodes image features locally. In the rightmost two cases, two \inr{}, trained with a shared decoder, are mixed by masking specific latents or by simply stacking them.}
    \label{fig:inr_transformations}
\end{figure}

We first consider transformations on the \inr{} coordinates. Given a trained \inr{}, we manipulate only the latent vector positions. The index function cache is refreshed and a forward pass is performed. In \autoref{fig:inr_transformations}, we perform several such transformations and show the resulting reconstructions. We can indeed verify that transforming latent vector positions yields correspondingly transformed reconstructions. If the anchored latents represent local image content, it is possible to mix \inr{}s. We can \eg select parts of different \inr{}s or simply stack them. This is shown on the right in \autoref{fig:inr_transformations}, where two jointly trained \inr{}s produce interesting reconstructions when mixed. In areas where the image gradient norm of either image is high, we retain the original image features. As such, mixing \inr{}s yields a reconstruction in which the original images are still relatively well represented. This finding inspired our \inr{}mix data augmentation method.

Let us investigate the local feature representations' significance on \inr{} classification accuracy. We reuse the FMNIST dataset that was padded to $100 \times 100$ in Experiment 1. The latent vectors of this dataset are pushed into a random direction once, and then used to 1) evaluate a normally trained PTv3 instance, and 2) train a PTv3 instance from scratch. In \autoref{tab:exp_coordinateshifts} we list the resulting test accuracies. Unsurprisingly, keeping the latent vector positions intact leads to the highest test accuracy. The regularly trained PTv3 instance shows a large drop in test accuracy when evaluated on the pushed data, as the learned relative positional encoding within PTv3 becomes meaningless. Conversely, when we train on pushed \inr{} data, the classifier still works relatively well. We hypothesise that the latent features are descriptive enough that, through pooling and attention, PTv3 learns a relatively informative global context. Accuracy drops when evaluated on intact \inr{}s, which we attribute to PTv3 learning inconsistent or incorrect relative positional encodings from the displaced data.

\begin{figure}
    \centering
    \includegraphics[width=0.5\linewidth]{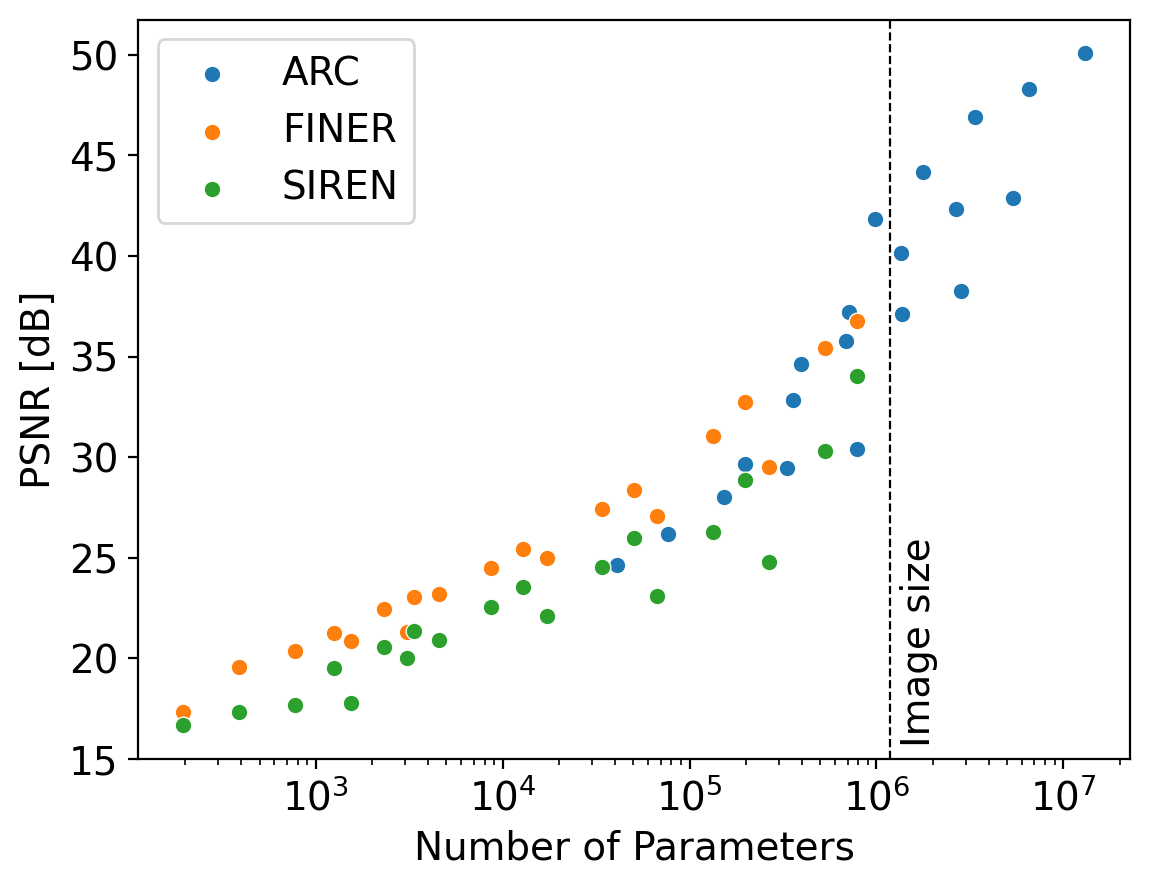}
    \caption{Image reconstruction results on the KODAK dataset. We compare \inr{} against SIREN and FINER, reporting PSNR as a function of model capacity. The vertical line represents the parameter count of a raw KODAK image, indicating the memory efficiency of different INR representations. SIREN and FINER experiments were bounded by computation costs, whereas \inr{} was not. \inr{} has a higher lower bound on memory than baselines but demonstrates superior scaling capabilities.}
    \label{fig:app_psnr}
\end{figure}

\paragraph{\inr{} reconstruction quality.}
Obtaining a high parameter efficiency or high reconstruction quality is not the primary objective of \inr{}. However, as it is commonplace in INR research, we will evaluate \inr{}'s reconstruction capabilities and compare it to several baselines. Since reconstruction quality scales with INR capacity, we investigate various configurations of each INR type and report the number of parameters alongside the obtained PSNR after 1000 training steps. Each configuration is used to fit the KODAK image dataset \citep{kodak} three times, taking the mean over all three runs and final reconstruction PSNRs. A mini-batch ratio of 1.0 is used except for \inr{} which uses a mini-batch ratio of 0.5. Results for SIREN, FINER \citep{liu2024finer}, and \inr{}, are shown in \autoref{fig:app_psnr}. For \inr{}, all combinations of $\{0.01, 0.05, 0.1, 0.25\} \cdot \text{\small \#pixels}$ latent vectors and latent dimensions $\{8,    16,   32,   64,   128\}$ are used. For SIREN and FINER, combinations of $\{1,2,3,4\}$ hidden layers and $\{ 32, 64, 128, 256, 512 \}$ hidden dimension are used. SIREN and FINER dimensions could not be increased further due to the computation time increasing very steeply since all their weights must be updated for each supervised coordinate. Contrastingly, \inr{} uses 4 neighbouring latents for each coordinate, regardless of its capacity. \inr{} is therefore significantly faster to train and more scalable.

A vertical line denotes the number of values a KODAK image contains if it were loaded into memory at $768\cdot512\cdot3=1.1\text{e}6$ parameters. Any INR instance with fewer parameters is arguably more memory-efficient than the original image. Larger INR representations show how PSNR increases if more memory is available, though such representations would not directly yield memory benefits. For \inr{}, the decoder's parameters are not included as the decoder is shared across all instances of a given architecture and its cost is thus amortised. \inr{} shows a generally higher parameter cost than MLP-based INR methods.

\subsection{Further ablations}
\paragraph{Ablation vs. classification accuracy.}
In this ablation experiment, we ablate various aspects of \inr{} and PTv3 to observe their impact on classification accuracy. For each ablation, \inr{}s are fit on a subset of 10k CIFAR10 images.

First, we compare latent vector normalisation techniques and their impact on validation accuracy. Three normalisation strategies are compared. None: where no normalisation is applied, Normalise Whole: where the latent vectors are normalised using a scalar mean and standard deviation which are precomputed on a subset of the \inr{}s, and Normalise Per-dim: where the mean and standard deviation are precomputed per latent dimension on a subset of the \inr{}s. The effect of these techniques on test accuracy is listed in \autoref{tab:normalisation_compar}. Generally, applying normalisation to the latent vectors improves classification accuracy. This is in line with findings in other INR classification literature \citep{dwsnets, zhou2024neural, zhou2024permutation, kofinas2024graph_nnsaregraphs}. Furthermore, applying normalisation across individual latent dimensions yields the highest improvement. This suggests that capturing variations specific to each latent dimension provides a more robust representation in classifying \inr{}s. We hypothesise that certain latent dimensions specialise in capturing distinct image features. This type of alignment would be induced by sharing the decoder. A similar property is introduced manually in \citet{chen2023neurbf} using harmonics of different frequencies, which results in better INR reconstructions.

Next, we explore the trade-off between the number of latent vectors and the latent dimensionality of each latent vector in \inr{}, and aim to analyse their impact on classification accuracy. We fit a subset of 10k CIFAR10 images to each combination. In \autoref{tab:latentdim_vs_numlatents} the validation accuracies which PTv3 converges to are listed. Accuracy improves significantly as we increase the number of latent vectors in the image. Conversely, increasing the latent dimension yields diminishing returns on accuracy.

\begin{table}[H]
    \begin{minipage}[b]{.47\linewidth}
        \centering
        \small
        \begin{tabular}{lc}
        \toprule
            \textbf{Latent normalisation} &  \textbf{Val. Acc.}\\
        \midrule
            None                &  52.67 \\    
            Normalise Whole     &  54.58 \\    
            Normalise Per-dim   &  58.68 \\    
        \bottomrule
        \end{tabular}
        \caption{Validation accuracy (\%$\uparrow$) under different latent normalisation strategies. We show how normalising each latent dimension independently yields the highest increase in performance.}
        \label{tab:normalisation_compar}
        
    \end{minipage} \hfill
    \begin{minipage}[b]{.47\linewidth}
        \small
        \centering
        \begin{tabular}{llccc}
        \toprule
                        & & \multicolumn{3}{c}{\textbf{Latent dimension}} \\
        \cmidrule{3-5}
                        & & \textbf{8}     & \textbf{16}    & \textbf{32}    \\
        \midrule
            \parbox[t]{2mm}{\multirow{3}{*}{\rotatebox[origin=c]{90}{\textbf{Latents}}}}
            & \textbf{10} & 30.95 & 32.91 & 29.01 \\
            & \textbf{25} & 41.12 & 40.05 & 41.43 \\
            & \textbf{50} & 49.14 & 48.62 & 48.02 \\
        \bottomrule
        \end{tabular}
        \caption{Validation accuracy (\%$\uparrow$) for differing number of latents and latent dimensions. Increasing the number of latents has a clear positive impact on classification accuracy, whereas high latent dimensions provide diminishing results, particularly when there are few latents.}
        \label{tab:latentdim_vs_numlatents}
    \end{minipage}
\end{table}

\begin{wrapfigure}{R}{0.35\textwidth}
    \centering
    \includegraphics[width=\linewidth]{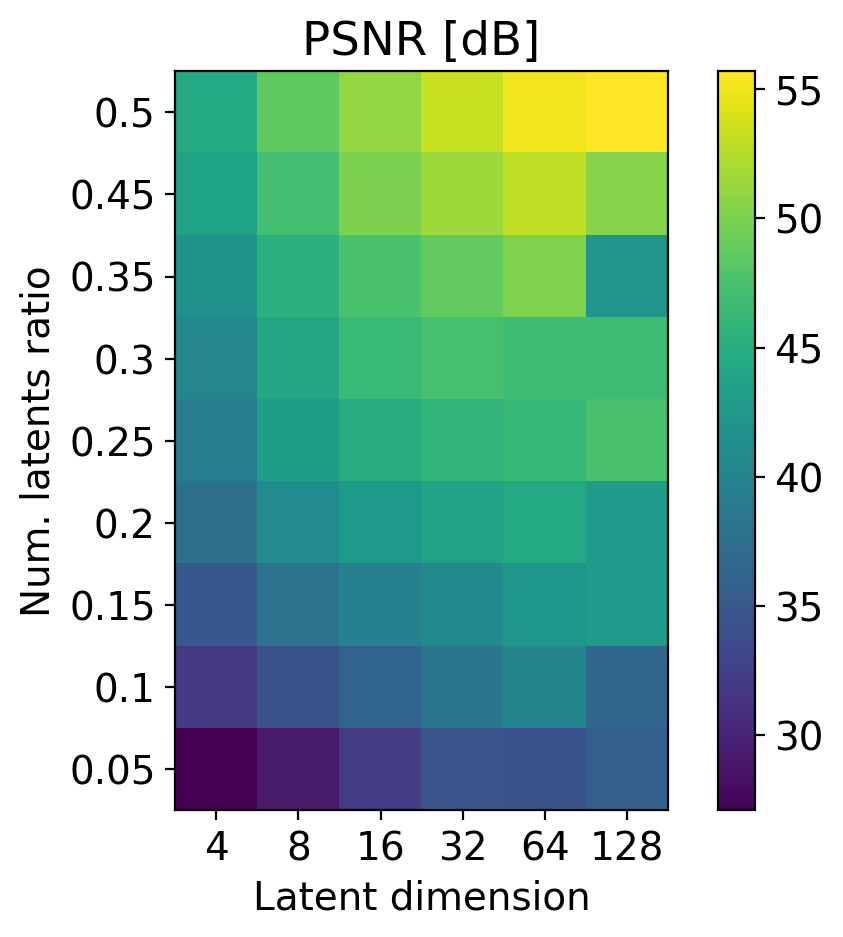}
    \caption{PSNR for different \inr{} configurations on skimage's astronaut image. The vertical axis denotes the ratio of the image's pixels that contain a latent vector.}
    \label{fig:app_heatmap_psnr}
\end{wrapfigure}

\paragraph{Ablation vs. PSNR.}
In this ablation experiment, we investigate different \inr{} configurations and observe their effect on image reconstruction quality. Given skimage's astronaut image, we train an \inr{} instance with various combinations of latent dimensions and number of latents, the latter being expressed as a factor of the number of pixels in the image. \autoref{fig:app_heatmap_psnr} shows the gradual increase in reconstruction quality as either the number of latents or the latent dimensions increase. This plot is generally in line with our expectations.

\subsection{Implementation details}

\paragraph{\inr{} implementation details.} \label{sec:app_inr_implementation}
Provided with an image, the latent positions are determined by sampling the image gradient norm. Like \citep{chen2023neurbf}, the number of latent vectors is proportional to the number of pixels in the image. We experimented with different combinations of the number of latents and latent dimension and found that $\text{\# latents}= 0.05 \cdot \text{\# image pixels}$, combined with a feature dimension of $z=32$ works well for most cases. This configuration is used in all experiments, unless otherwise noted. Like \citep{chen2023neurbf}, the feature vectors are drawn from $ U(-1\mathrm{e}{-4}, 1\mathrm{e}{-4})$. Upon initialisation, the indexing function $U_n$ caches the index and relative position to the $n$ nearest latent vectors. In all our experiments, we use a decoder of size $[n\cdot (z+2), n\cdot (z+2), d_{\text{out}} ]$. The decoder is pre-trained by jointly training 100 \inr{} instances, aggregating the loss over all instance for each step. All our \inr{} instances, as well as the pretrained decoder, are trained for 500 steps. In fitting \inr{} we make use of mini-batching, whereby only $25 \%$ of the image coordinates are supervised each iteration. We found that this makes qualitatively little difference in reconstruction quality but makes fitting significantly faster. An ADAM optimiser is used with a learning rate of $0.005$. We normalise images to [0,1] range.

\paragraph{Point Transformer v3 implementation details.}
Point Transformer v3 (PTv3) \citep{wu2024ptv3} is not intended to be used outside of point-cloud tasks, which typically contain a 3D coordinate with optional features such as colours or normals. Hence, PTv3 requires a few tweaks in order to be compatible with \inr{}. For instance, the latent coordinates are made three-dimensional by appending a 0 to them. To make PTv3 suitable for classification tasks, we replace the standard upsampling blocks by a global pooling layer, followed by a single linear layer that maps the feature dimension to the number of classes.

\begin{figure}
    \centering
    \begin{subfigure}{0.2\linewidth}
        \footnotesize
        \centering
        Ground Truth
        \includegraphics[width=\textwidth]{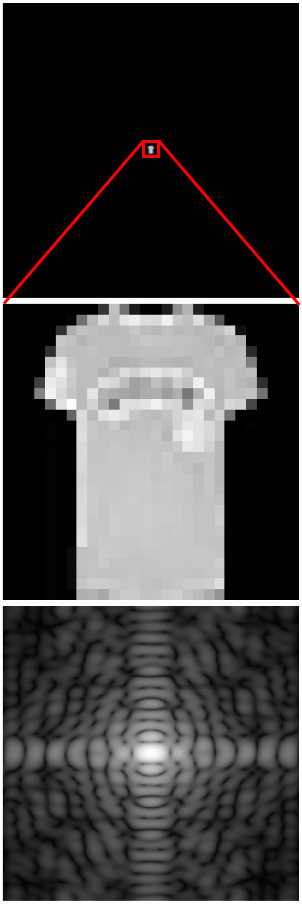}
    \end{subfigure}
    \begin{subfigure}{0.2\linewidth}
        \footnotesize
        \centering
        DWSnets SIREN
        \includegraphics[width=\textwidth]{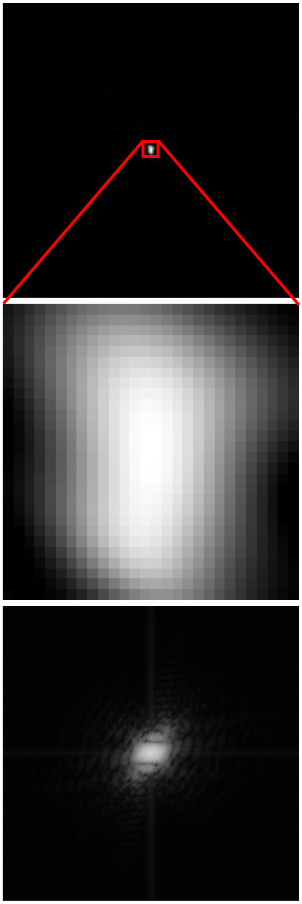}
    \end{subfigure}
    \begin{subfigure}{0.2\linewidth}
        \footnotesize
        \centering
        Standard SIREN
        \includegraphics[width=\textwidth]{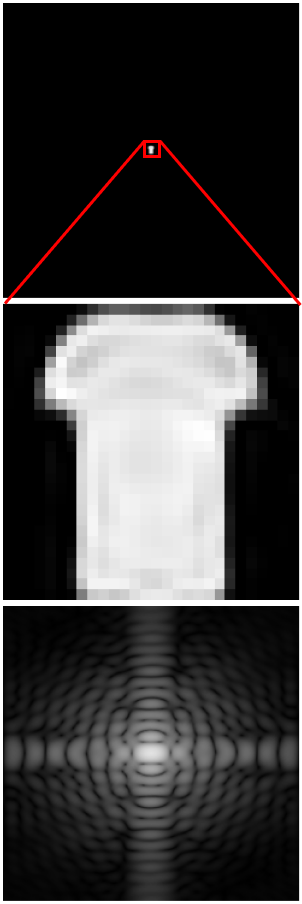}
    \end{subfigure}
    \caption{Comparison of $1024 \times 1024$ FMNIST objects reconstructed using DWSnets' SIREN implementation and a corrected SIREN implementation. Rows depict (from top to bottom) the full image, a zoomed-in view, and the FFT of the reconstruction. The left column shows the original image, highlighting its increased frequency content. From the FFTs, it is clear that spectral bias phenomena re-emerge in the SIREN INRs.}
    \label{fig:app_dwsnets_blur}
\end{figure}

\paragraph{SIREN implementation details.} \label{sec:app_dwsnets}
Several of our experiments required custom SIREN datasets. We aimed to follow DWSnets implementation but found it to be erroneous compared to the regular SIREN specification \citep{sitzmann2019siren}. Specifically, a $0.5$ offset is added to the network's output, the first layer does not follow the prescribed initialisation scheme, and the bias layer is initialised to zero rather than the prescribed weight initialisation. We found that these errors are not readily apparent in low-resolution images such as the ones commonly used in INR classification. In fitting our larger resolution images, we observe heavy blurring in the reconstruction \autoref{fig:app_dwsnets_blur}. We opted to use a correct SIREN implementation instead.

We use default SIREN hyperparameters; $\omega_0=30.0$, no final layer activation function, ADAM optimiser and learning rate of $0.0005$. The SIRENs are trained for 1000 steps.

\subsection{Baseline details} \label{sec:app_expdetails}
\paragraph{ScaleGMN.} In utilising the ScaleGMN baseline \citep{kalogeropoulos2024scaleequivariant}, it proved to be rather unstable in training. The authors acknowledge this issue and take measures such as layer normalisation and skip connections to mitigate it. ScaleGMN is designed to work on SIREN datasets introduced by the DWSnets paper \citep{dwsnets}. We found that a faulty SIREN implementation was used in creating these datasets (\autoref{sec:app_dwsnets}). The SIREN datasets that we created therefore present an extra challenge as the provided ScaleGMN settings were created with DWSnet-SIRENs in mind. To quantify this error, we train ScaleGMN using the provided MNIST settings on 10k SIRENs that we fit ourselves. The resulting test accuracy is $88.89\%$ after 71 epochs, which is $7.68 \%$ lower than the test accuracy which the authors obtained $96.57 \%$ on the whole MNIST dataset. We proceed to use ScaleGMN in our experiments where we use our own SIREN datasets, but are aware of the possibly suboptimal performance.

\paragraph{Toy datasets.} In most experiments with toy datasets FMNIST was chosen as a basis. FMNIST poses a non-trivial classification challenge for current methods \citep{dwsnets, kofinas2024graph_nnsaregraphs, kalogeropoulos2024scaleequivariant} and can easily be padded to increase the image size.

\begin{figure}
\centering
    \includegraphics[width=0.5\linewidth]{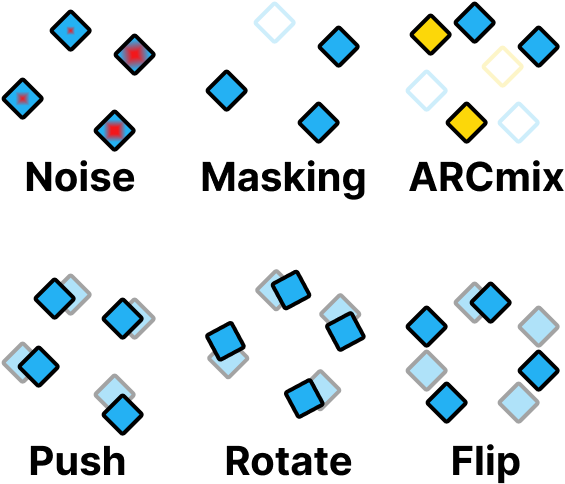}
    \caption{Data augmentation techniques for \inr{}s. Noise, Masking, and \inr{}mix manipulate the latent cloud to enhance data diversity, while Push, Rotate, and Flip change just the latent vector coordinates. These augmentations operate directly in the \inr{} weight-space.}
    \label{fig:data_augmentation}
\end{figure}

\begin{figure}
    \centering
    \includegraphics[width=0.5\linewidth]{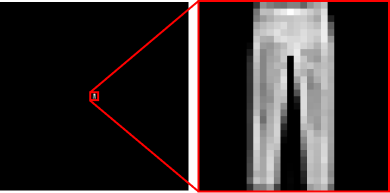}
    \caption{Left, a sample from the $1024 \times 1024$ padded FMNIST dataset. Right, a zoomed in view of the depicted object. The high-resolution of the image, transforms the relatively simple task of FMNIST classification into a challenge for baseline INR classification methods. }
    \label{fig:app_highres_preview}
\end{figure}

\begin{figure}
    \centering
    \begin{subfigure}{0.3\linewidth}
        \includegraphics[width=\textwidth]{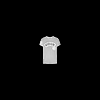}
        \centering
        \footnotesize
        \centeredZero{}
    \end{subfigure}
    \begin{subfigure}{0.3\linewidth}
        \includegraphics[width=\textwidth]{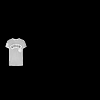}
        \centering
        \footnotesize
        \displacedZero{}
        \label{fig:exp_transinvar_zeropadded}
    \end{subfigure}
    \captionsetup{type=figure}
    \caption{\textbf{Exp. 5:} Samples from the \centeredZero{} and \displacedZero{} datasets.}
    \label{fig:transinvar_0s}
\end{figure}

\end{document}